\documentclass[twoside,leqno,twocolumn]{article}

\usepackage[letterpaper]{geometry}

\usepackage{ltexpprt}
\usepackage{hyperref}

\usepackage[title]{appendix}
\usepackage{bbold}
\usepackage{algorithm}
\usepackage{algorithmic}
\usepackage{amssymb}
\usepackage{array}
\usepackage{booktabs}
\usepackage{enumitem}

\usepackage{microtype}
\usepackage{graphicx}
\usepackage{booktabs} 
\usepackage{hyperref}
\usepackage{wrapfig}
\usepackage{amsmath}
\usepackage{subfig}

\usepackage{balance} 
\usepackage{setspace}
\usepackage[T1]{fontenc}

\usepackage{makecell}

\newtheorem{definition}{Definition}[section]
\usepackage{color}

\begin{document}

\newcommand\relatedversion{}
\renewcommand\relatedversion{\thanks{The full version of the paper can be accessed at \protect\url{https://arxiv.org/abs/1902.09310}}} % Replace URL with link to full paper or comment out this line

\title{Enhancing Adversarial Training with Feature Separability}
\author{Yaxin Li
\and Xiaorui Liu
\and Han Xu
\and Wentao Wang
\and Jiliang Tang
\thanks{Michigan State University. \{liyaxin1, xiaorui, xuhan1, wangw116\}@msu.edu}
}

% \author{
% \IEEEauthorblockN{Yaxin Li}
% \IEEEauthorblockA{\textit{Dept. of Comp. Sci. and Engr.} \\
% \textit{Michigan State University}\\
% East Lansing, MI \\
% liyaxin1@msu.edu}
% \and
% \IEEEauthorblockN{Xiaorui Liu}
% \IEEEauthorblockA{\textit{Dept. of Comp. Sci. and Engr.} \\
% \textit{Michigan State University}\\
% East Lansing, MI \\
% xiaorui@msu.edu}
% \and
% \IEEEauthorblockN{Han Xu}
% \IEEEauthorblockA{\textit{Dept. of Comp. Sci. and Engr.} \\
% \textit{Michigan State University}\\
% East Lansing, MI \\
% xuhan1@msu.edu}
% \linebreakand

% \and
% \IEEEauthorblockN{Wentao Wang}
% \IEEEauthorblockA{\textit{Dept. of Comp. Sci. and Engr.} \\
% \textit{Michigan State University}\\
% East Lansing, MI \\
% wangw116@msu.edu}
% \and
% \IEEEauthorblockN{Jiliang Tang}
% \IEEEauthorblockA{\textit{Dept. of Comp. Sci. and Engr.} \\
% \textit{Michigan State University}\\
% East Lansing, MI \\
% tangjili@msu.edu}
% }

\date{}

\maketitle

% Copyright Statement
% When submitting your final paper to a SIAM proceedings, it is requested that you include
% the appropriate copyright in the footer of the paper.  The copyright added should be
% consistent with the copyright selected on the copyright form submitted with the paper.
% Please note that "20XX" should be changed to the year of the meeting.

% Default Copyright Statement
\fancyfoot[R]{\scriptsize{Copyright \textcopyright\ 20XX by SIAM\\
Unauthorized reproduction of this article is prohibited}}

% Depending on which copyright you agree to when you sign the copyright form, the copyright
% can be changed to one of the following after commenting out the default copyright statement
% above.

%\fancyfoot[R]{\scriptsize{Copyright \textcopyright\ 20XX\\
%Copyright for this paper is retained by authors}}

%\fancyfoot[R]{\scriptsize{Copyright \textcopyright\ 20XX\\
%Copyright retained by principal author's organization}}

%\pagenumbering{arabic}
%\setcounter{page}{1}%Leave this line commented out.

\begin{abstract}
Deep Neural Network (DNN) are vulnerable to adversarial attacks. As a countermeasure, adversarial training aims to achieve robustness based on the min-max optimization problem and it has shown to be one of the most effective defense strategies. However, in this work, we found that compared with natural training, adversarial training fails to learn better feature representations for either clean or adversarial samples, which can be one reason why adversarial training tends to have severe overfitting issues and less satisfied generalize performance.
Specifically, we observe two major shortcomings of the features learned by existing adversarial training methods:(1) low intra-class feature similarity; and (2) conservative inter-classes feature variance. 
To overcome these shortcomings, we introduce a new concept of adversarial training graph (ATG) with which the proposed adversarial training with feature separability (ATFS) enables to coherently boost the intra-class feature similarity and increase inter-class feature variance. Through comprehensive experiments, we demonstrate that the proposed ATFS framework significantly improves both clean and robust performance. The implementation of this work can be found in the link\footnote{\url{https://github.com/no-name-submission/ATFS}}.
\end{abstract}

\newcommand{\va}{{\mathbf{a}}}
\newcommand{\vb}{{\mathbf{b}}}
\newcommand{\vc}{{\mathbf{c}}}
\newcommand{\vd}{{\mathbf{d}}}
\newcommand{\ve}{{\mathbf{e}}}
\newcommand{\vf}{{\mathbf{f}}}
\newcommand{\vg}{{\mathbf{g}}}
\newcommand{\vh}{{\mathbf{h}}}
\newcommand{\vi}{{\mathbf{i}}}
\newcommand{\vj}{{\mathbf{j}}}
\newcommand{\vk}{{\mathbf{k}}}
\newcommand{\vl}{{\mathbf{l}}}
\newcommand{\vm}{{\mathbf{m}}}
\newcommand{\vn}{{\mathbf{n}}}
\newcommand{\vo}{{\mathbf{o}}}
\newcommand{\vp}{{\mathbf{p}}}
\newcommand{\vq}{{\mathbf{q}}}
\newcommand{\vr}{{\mathbf{r}}}
\newcommand{\vs}{{\mathbf{s}}}
\newcommand{\vt}{{\mathbf{t}}}
\newcommand{\vu}{{\mathbf{u}}}
\newcommand{\bv}{{\mathbf{v}}}
\newcommand{\vw}{{\mathbf{w}}}
\newcommand{\vx}{{\mathbf{x}}}
\newcommand{\vy}{{\mathbf{y}}}
\newcommand{\vz}{{\mathbf{z}}}

\newcommand{\vA}{{\mathbf{A}}}
\newcommand{\vB}{{\mathbf{B}}}
\newcommand{\vC}{{\mathbf{C}}}
\newcommand{\vD}{{\mathbf{D}}}
\newcommand{\vE}{{\mathbf{E}}}
\newcommand{\vF}{{\mathbf{F}}}
\newcommand{\vG}{{\mathbf{G}}}
\newcommand{\vH}{{\mathbf{H}}}
\newcommand{\vI}{{\mathbf{I}}}
\newcommand{\vJ}{{\mathbf{J}}}
\newcommand{\vK}{{\mathbf{K}}}
\newcommand{\vL}{{\mathbf{L}}}
\newcommand{\vM}{{\mathbf{M}}}
\newcommand{\vN}{{\mathbf{N}}}
\newcommand{\vO}{{\mathbf{O}}}
\newcommand{\vP}{{\mathbf{P}}}
\newcommand{\vQ}{{\mathbf{Q}}}
\newcommand{\vR}{{\mathbf{R}}}
\newcommand{\vS}{{\mathbf{S}}}
\newcommand{\vT}{{\mathbf{T}}}
\newcommand{\vU}{{\mathbf{U}}}
\newcommand{\vV}{{\mathbf{V}}}
\newcommand{\vW}{{\mathbf{W}}}
\newcommand{\vX}{{\mathbf{X}}}
\newcommand{\vY}{{\mathbf{Y}}}
\newcommand{\vZ}{{\mathbf{Z}}}

\newcommand{\cA}{{\mathcal{A}}}
\newcommand{\cB}{{\mathcal{B}}}
\newcommand{\cC}{{\mathcal{C}}}
\newcommand{\cD}{{\mathcal{D}}}
\newcommand{\cE}{{\mathcal{E}}}
\newcommand{\cF}{{\mathcal{F}}}
\newcommand{\cG}{{\mathcal{G}}}
\newcommand{\cH}{{\mathcal{H}}}
\newcommand{\cI}{{\mathcal{I}}}
\newcommand{\cJ}{{\mathcal{J}}}
\newcommand{\cK}{{\mathcal{K}}}
\newcommand{\cL}{{\mathcal{L}}}
\newcommand{\cM}{{\mathcal{M}}}
\newcommand{\cN}{{\mathcal{N}}}
\newcommand{\cO}{{\mathcal{O}}}
\newcommand{\cP}{{\mathcal{P}}}
\newcommand{\cQ}{{\mathcal{Q}}}
\newcommand{\cR}{{\mathcal{R}}}
\newcommand{\cS}{{\mathcal{S}}}
\newcommand{\cT}{{\mathcal{T}}}
\newcommand{\cU}{{\mathcal{U}}}
\newcommand{\cV}{{\mathcal{V}}}
\newcommand{\cW}{{\mathcal{W}}}
\newcommand{\cX}{{\mathcal{X}}}
\newcommand{\cY}{{\mathcal{Y}}}
\newcommand{\cZ}{{\mathcal{Z}}}

\newcommand{\norm}[1]{\left\lVert#1\right\rVert}

\section{Introduction}
\label{sec:intro}

% \xr{
% logic: mainly follow the abstract. 

% 1. necessary backgound and adv train is the best solution so far.

% 2. However, recent works demonstrate that adv train tends to overfit and lots of defense strategy cannot beat basic adv train with early stopping. Therefore, to further improve adv train is crucial. 

% 3. We found that adv train doesn't help the model to learn a better feature representation either for clean or adv samples which might be one reason for robust overfitting (show our preliminary result). 

% 4. Summarize the shortcome of adv train we observe. 

% 5. This motivates us to enahnce adv train with feature separability and what are the challenges; 

% 6 summarize our contributions }

%Deep learning has been applied in many safety critical applications such as auto driving~\cite{bojarski2016end} and healthcare~\cite{esteva2019guide}. However,  adversarial attacks~\cite{goodfellow2014explaining}.

%\wt{it's better to unify clean/adversarial samples OR clean/adversarial examples throughout the whole paper}
Although deep neural networks (DNNs) have achieved extraordinary accomplishments on various machine learning tasks, their vulnerability to adversarial attacks ~\cite{goodfellow2014explaining, carlini2017towards} still raise great concerns when they are adopted to safety-critical tasks, such as autonomous vehicles~\cite{bojarski2016end} and AI healthcare~\cite{esteva2019guide}.
As one of the most effective and reliable countermeasures to protect DNNs against adversarial attacks, adversarial training methods~\cite{madry2017towards, goodfellow2014explaining, zhang2019theoretically} train DNNs to correctly classify the manually generated adversarial examples around the clean samples. Particularly, they optimize the model to have the minimum adversarial risk of a perturbed adversarial sample.
%\jt{ I think we need to remove this formulation since we did not introduce anything about it and people will confuse by it}
% \begin{align}
%     \min_f \underset{x}{\E}~  \left[\max_{||\delta||\leq\epsilon} \mathcal{L}(f(x+\delta),y)\right].
%     \label{eq:adv_training}
% \end{align}
%
%Although Adversarial Training have been shown to be effective, the performance is still unsatisfactory. One one hand, clean performance is seriously reduced compares to clean training; On the other hand, due to robust generalization gap, test robust accuracy is significantly worse than training robust accuracy, thus lead to low robust performance. 
%
%
% due to the large generalization gap. 

However, adversarial training methods still suffer from major drawbacks in practice. For example, the adversarially trained models always have lower clean accuracy than naturally trained models~\cite{tsipras2018robustness}. Furthermore, recent works~\cite{schmidt2018adversarially, wu2020adversarial, rice2020overfitting} show that adversarially trained models have strong overfitting issues for adversarial robustness. In order to overcome these issues and improve the performance of adversarially trained models, some existing works attempt to leverage a much larger training set~\cite{schmidt2018adversarially} or larger model architectures~\cite{katz2017towards}. While, how to enhance adversarial training from algorithmic perspectives is still an open problem and requires deeper exploration.

%Some strategies try to introduce more training data~\cite{schmidt2018adversarially} and larger capacity model structures~\cite{nakkiran2019adversarial} .
%With limited data and model capacity, it is challenging to learn the adversarial examples that mix up around the decision boundary~\cite{zhang2020geometry}. In order to mitigate this overfitting issue, efforts have been made to focus on more valuable data~\cite{wang2019improving}, and study the weight loss landscape~\cite{wu2020adversarial}.
% \iffalse
% \xr{With insufficient data and capacity, some proposed methods tend to focus on more valuable data~\cite{wang2019improving} ~\cite{zhang2020geometry} or directly design weight perturbation to increase model generalization ~\cite{wu2020adversarial}.
% }\fi

\begin{figure*}[t]
\subfloat[Natural Training]{\label{clean}
\begin{minipage}[c]{0.25\textwidth}
\centering
\includegraphics[width = 1\textwidth]{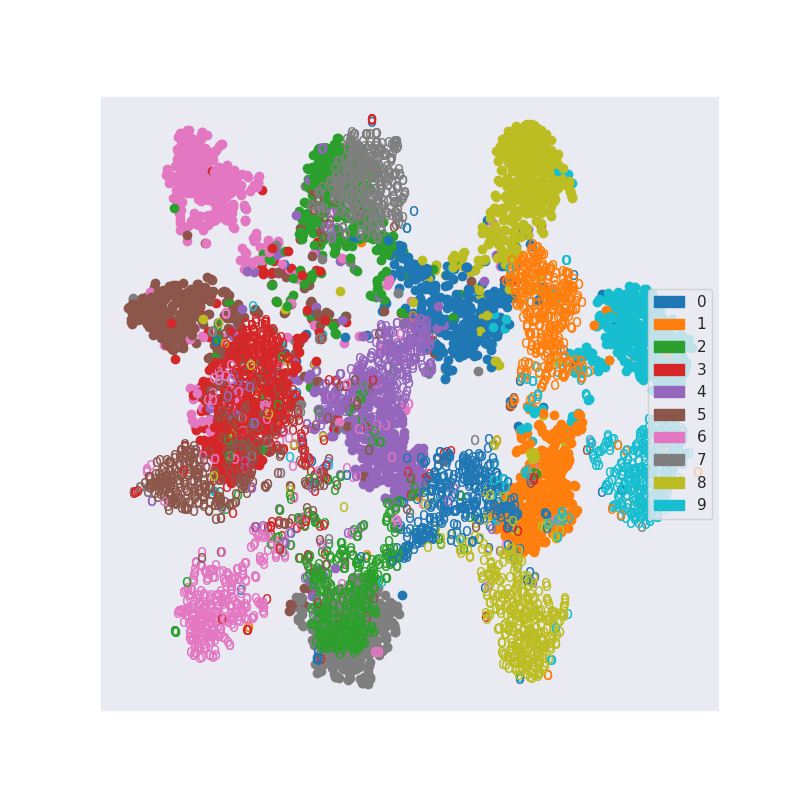}
\end{minipage}
}
%\subfloat[PGD Training (2/255)]{ \label{pgd_2}
%\begin{minipage}[c]{0.24\textwidth}
%\centering
%\includegraphics[width = 1\textwidth]{pic/tsne_pgd_2.png}
%\end{minipage}
%}
%\subfloat[PGD Training (4/255)]{  \label{pgd_4}
%\begin{minipage}[c]{0.26\textwidth}
%\centering
%\includegraphics[width = 1\textwidth]{pic/tsne_pgd_4.png}
%\end{minipage}
%}
\subfloat[PGD Training]{\label{pgd}
\begin{minipage}[c]{0.25\textwidth}
\centering
\includegraphics[width = 1\textwidth]{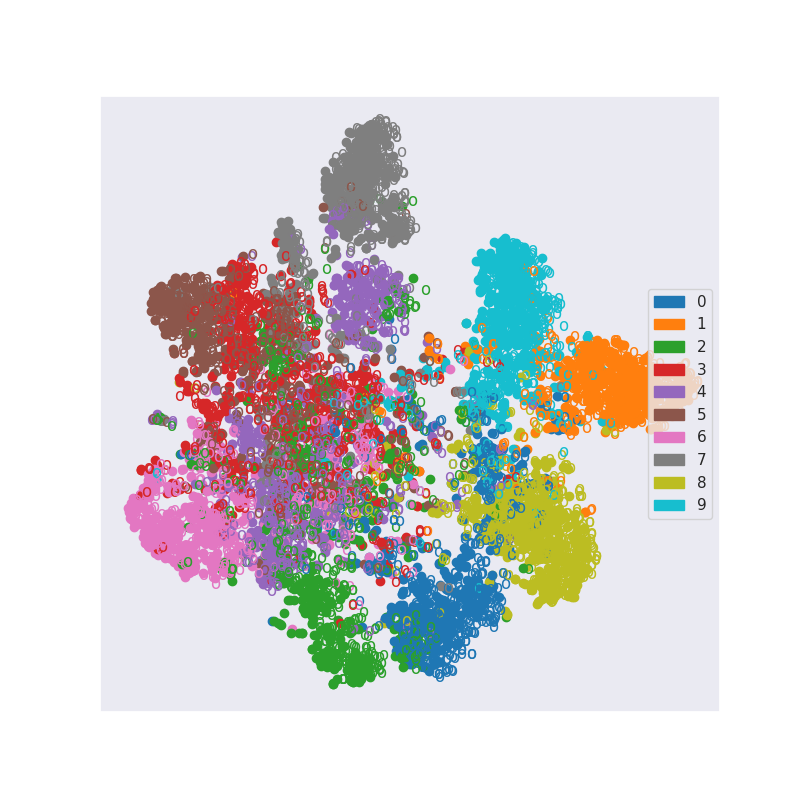}
\end{minipage}
}
\subfloat[TRADES]{\label{trade}
\begin{minipage}[c]{0.25\textwidth}
\centering
\includegraphics[width = 1\textwidth]{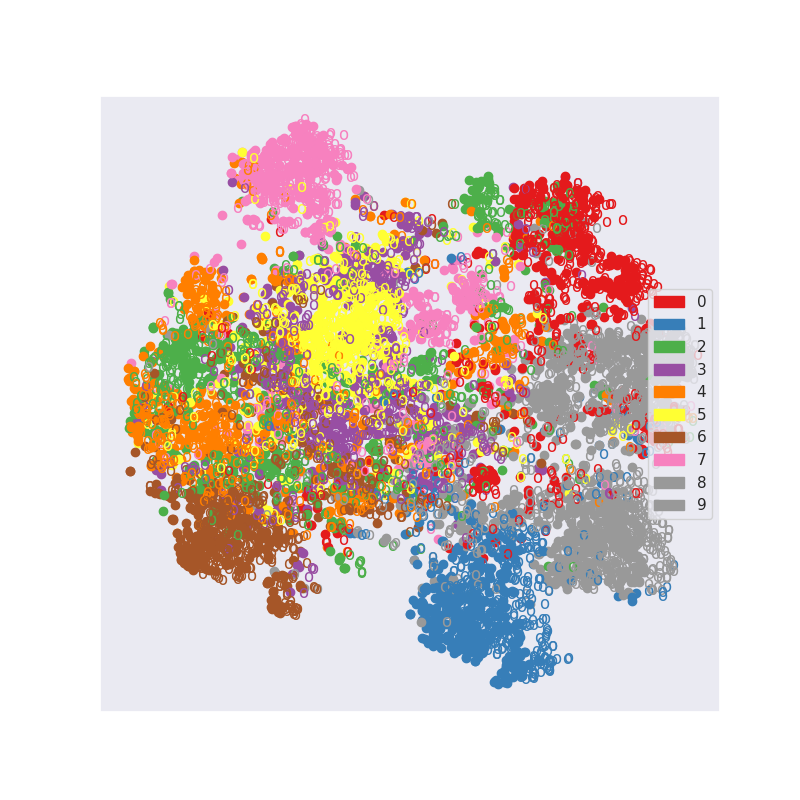}
\end{minipage}
}
\subfloat[MART]{\label{awp}
\begin{minipage}[c]{0.25\textwidth}
\centering
\includegraphics[width = 1\textwidth]{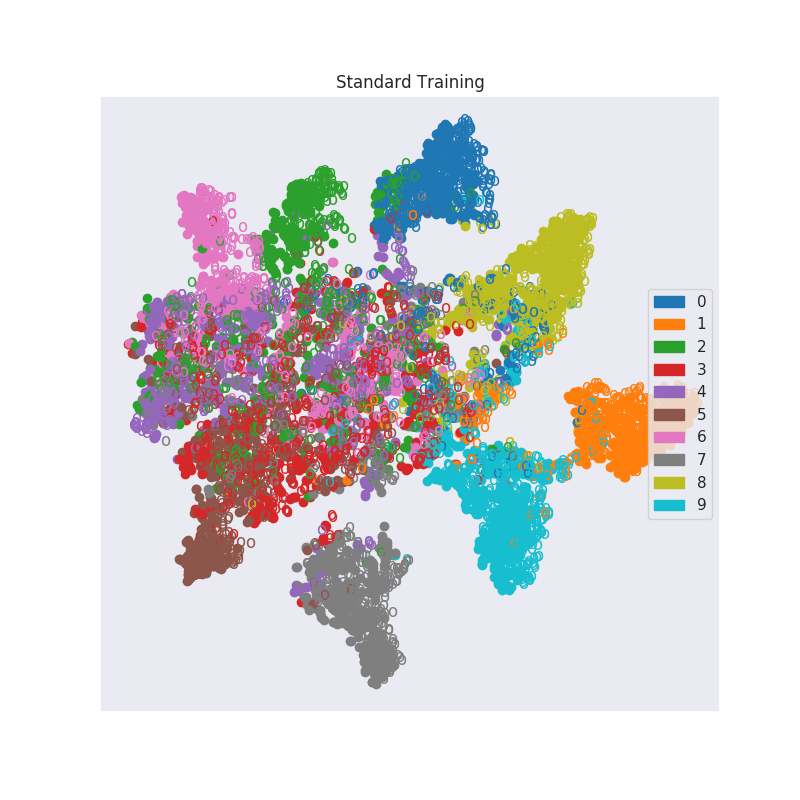}
\end{minipage}
}
\caption{t-SNE feature visualization  of naturally and adversarially trained models. The solid dots denote clean test samples and the hollow dots represent adversarial test samples generated by PGD-20 (8/255).}
\label{fig:Visualization}
\end{figure*}

In this work, we investigate potential reasons of the overfitting issues of adversarial training by examining the quality of their learned features.
% for both clean and adversarial samples.
In our preliminary study, we observe that adversarial trained models always learn ``unsatisfactory'' features for both clean and adversarial samples. 
% In particular, we found two shortcomings of existing adversarial training methods: (1) low intra-class feature similarity: feature similarities for samples from the same class are low and their is no clear cluster structures for each class; 
% and (2) conservative inter-classes feature variance: the distance between features from different classes are small and there is no no clear separability. 
% These shortcomings appear to be the potential reasons why adversarial training can not achieve both superior clean and robustness performance.
As an illustration, Figure \ref{fig:Visualization} shows the learned features extracted from the second last layer of ResNet18~\cite{he2016deep} models, which are trained by different algorithms on CIFAR10~\cite{krizhevsky2009learning} datasets, including natural training, PGD adversarial training~\cite{madry2017towards}, TRADES~\cite{zhang2019theoretically} and the AWP method~\cite{wu2020adversarial}. The extracted features are visualized in a 2-dimensional space by t-SNE~\cite{van2008visualizing}. 
% on the CIFAR10 dataset where Figure~\ref{clean} for a naturally trained model while Figure ~\ref{pgd_2}, ~\ref{pgd_4}, ~\ref{pgd_8} for adversarially trained models with perturbation budget of $2/255$, $4/255$ and $8/255$, separately. 
From Figure~\ref{fig:Visualization}, we can make the following observations:
\begin{enumerate}[leftmargin=*]
    \item A naturally trained model can learn discriminative features for clean samples. Samples from different classes are well separated and the features of samples from the same class are well clustered. However, a naturally trained model cannot learn good features for the adversarial samples and the features of adversarial and clean samples are far away from each other. This can reflect the reason why naturally trained models can achieve high clean generalization but low robust performance. 
    \item For all adversarially trained models, including PGD training, TRADES and MART, the features of samples have much poorer ``quality'' than natural training. For example, for both clean or adversarial examples, \textbf{(a)} samples from the same class are not clustered together in the feature space;  and \textbf{(b)} samples from different classes are not well separated. As a result, the trained models have poor ability to discriminate samples from different classes via the learned features. Thus, it could be one key reason why the adversarially trained models suffer from both unsatisfactory clean accuracy \& adversarial robustness.
\end{enumerate}

These observations reveal two shortcomings of existing adversarial training methods. First, the inter-class feature variance is conservative. In other words, the model tends to embed samples from different classes with similar features. This makes the samples from different classes indistinguishable from each other and therefore degrades the clean classification performance and also contributes to unsatisfactory robustness. 
Second, low intra-class feature similarity results in a lack of clear clusters in the feature space. Therefore, the model can not easily extract the pattern for a specific class, which can degrade the clean performance.
These shortcomings provide possible explanations for the phenomenon that adversarial training tends to have a large natural and robust generalization gap and can not achieve both high clean and robust performance. %This is consistent with the finding that the cross-over mixture property of the adversarial samples will lead to the overfitting issue~\cite{zhang2020attacks}.

% The above preliminary study motivate us 
Motivated by these findings, we aim to enhance adversarial training by mitigating the aforementioned shortcomings. We first introduce a novel concept of adversarial training graph (ATG) that can help encode various types of relations among clean and adversarial samples in adversarial training. By capturing the link information in ATG, the proposed adversarial training framework, ATFS, is facilitated to calibrate the learned features to render desired feature properties, including: (1) increasing intra-class feature similarity of both clean \& adversarial examples, (2) increasing inter-class feature dissimilarity of both clean \& adversarial examples, and (3) increasing feature similarity between each clean sample and its adversarial counterpart. Experimental results on benchmark datasets show that ATFS can achieve better performance than a variety of representative baselines. We also conduct experiments to illustrate how ATFS mitigates the aforementioned shortcomings by feature visualization and boundary evaluation.    

%
%and improve the feature similarity within class and separate the features from different classes, we need to not only utilize the knowledge provided by label, but also the relationship between classes. These relationship between samples from different class can be viewed as a graph, where each node represent an input data and the link represent the relation of two different samples. Two samples from the same class is connected with a positive link. By maximizing the node similarity of representations where there is a positive link and minimizing the node similarity where the positive link is absent, the model makes full use of the information contained in the data and learns better representations. 
%
%Our contributions are listed as follows: 
%\begin{itemize}
%    \item We conduct extensively experiments shows that proposed method is effective to reduce robust generalization and achieve better performance than benchmarks.
 %   \item We illustrate that proposed method could help feature invariant among different classes and feature similarity within the same class. 
  %  \item We carefully study how those components could help %adversarial training. 
%\end{itemize}

The remaining of the paper is organized as follows: 
in Section~\ref{sec:methodology}, we introduce the concept of ATG and the proposed ATFS framework in detail; in Section~\ref{sec:experiment}, 
% we show the experimental results to demonstrate the effectiveness of our model from different perspectives
we demonstrate the effectiveness of our model through empirical comparison with the state-of-art baselines and further ablation study;
in Section~\ref{sec:related}, some related works are introduced; 
and in Section~\ref{sec:con}, we conclude this work and discuss future works.

% \iffalse
% \xr{Notes: 1. challenges; 2. emphasize non-trivial contribution. Can we formulate a unified framework for supervised CE training, adv training, TRADES from the perspective of graph. For supervised loss, we create a virtual node for each class and the prediction of link between each sample and the corresponding virtual node is equivalent to CE training (virtual node from other classes are negative nodes)}

% \xr{how about the visualization of normal cross-entropy training? maybe also compare to it to show how adv train destroy the feature in normal training. Also, don't need to show too many samples, show less samples will make the figure looks nicer}
% \fi
\section{The Proposed Framework}
\label{sec:methodology}

\subsection{Notations} Before we detail the proposed framework, we first introduce the necessary notations. In this paper, we consider the model is trained on the training set $\cD=\{(\vx_i, y_i)\}_{i=1}^{n}$, where $\vx_i$ is a clean training sample and $y_i\in \{1,...,C\}$ is the corresponding label with $C$ being the number of classes. Correspondingly, $\{\vx_i'\}_{i=1}^n$ represents the adversarial sample set generated from $\{\vx_i\}_{i=1}^n$ by the current model $f_\theta$. Note that an adversarial sample $\vx_i'$ is considered to have the same label $y_i$ as its clean sample $\vx_i$. 
As aforementioned, the adversarial training method and its variants minimize the model's loss on the adversarial sample $\vx'$ which is computed by the inner maximization problem, which can be formulated as:
\begin{align}
        \min_{f_\theta} \max_{\norm{\vx'-\vx}_{p} \leq \epsilon}\mathcal{L}(f_\theta(\vx'), y).
\end{align}
% \jt{we must explain this formulation and also notations? we should form the habit no matter what we want to use, let us first define it.}
%As a result, adversarial training tends to mix the adversarial samples with the samples from other classes in the feature space through the inner maximization process. Since adversarial training only trains the model on adversarial data, it is likely to overfit to the adversarial samples and give conservative prediction due to a vague decision boundary in the feature space. 
% \xr{the logic in above transition is not smooth}
Based on the discussion in Section~\ref{sec:intro}, we found that adversarial training algorithms always tend to learn mixed features. Motivated by this finding, we propose the \textit{Adversarial Training with Feature Separability} (ATFS) framework to calibrate the learned features for data samples that help achieve three major goals:
%(1) enhancing the robustness to adversarial samples; (2) boosting intra-class feature similarity; and (3) increasing inter-class feature variance. 
\begin{enumerate}
    \item Increase intra-class feature similarity of both clean \& adversarial examples, 
    \item Increase inter-class feature dissimilarity of both clean \& adversarial examples
    \item Increase feature similarity between each clean sample and its adversarial counterpart.
\end{enumerate}
To achieve these goals, we first introduce {\it Adversarial Training Graph (ATG)} to encode the desired relations between training samples. %ATG not only provides a comprehensive representation to denote relations among class labels, clean samples and adversarial samples, but also paves us a way to achieve these goals coherently by modeling relations in ATG. 
%Facilitated by ATG, the proposed ATFS can achieve ... 
In the following subsections, we will first introduce how to construct ATG for adversarial training. Then we detail the model components to capture different types of relations in ATG and consequently achieve the aforementioned goals. Finally, we present the final optimization objective and the proposed framework ATFS. 

% Before that, we first introduce the key notations we will use in the remaining of the paper.

\subsection{Adversarial Training Graph}
\label{sub: Adversarial_Training_Graph}
% \noindent \textbf{Notations.}

We formally define Adversarial Training Graph (ATG) as follows:

\begin{definition}[Adversarial training graph]
An adversarial training graph $\cG = \{\cV, \cE^+, \cE^-\}$ is an undirected, weighted and sign graph with the node set 
% $\cV=\{\vx_i\}_{i=1}^n \cup \{\vx_i'\}_{i=1}^n = \{\vx_i\}_{i=1}^{2n} $. 
$\cV=\{\vx_i\}_{i=1}^n \cup \{\vx_i'\}_{i=1}^n$. 
ATG contains a set of positive links $\cE^+$ and a set of negative links $\cE^-$ where $\cE^+$ connects nodes from the same class while $\cE^-$ connects nodes from different classes.
% $\cE^+=\{(\vx_i)\}$
% The node set $\cV$ contains two types of nodes including clean samples $\{\vx_i\}_{i=1}^{n}$ and adversarial samples $\{\vx_i^\prime\}_{i=1}^{n}$, i.e., $\cV = \{\vx_i, \vx_i^\prime\}_{i=1}^{n}$. 
% Note that the adversarial nodes are dynamically changing depending on current model.
\end{definition}

In ATG, a positive link indicates that its two ending nodes are desired to be similar; while a negative link encourages dissimilarity between two nodes. In particular, to achieve the aforementioned three goals, we introduce three types of links in ATG as follows:
\begin{itemize}[leftmargin=*]
  \setlength\itemsep{-0.2em}
    \item \textbf{Positive links $\cE^+$}. All samples from the same class as sample $\vx_i$ are connected with $\vx_i$ by a positive link. In particular, we construct two types of positive links:
    
    \textbf{1) $\cE_{\text{ca}}^+$}: To enhance the robustness against adversarial samples, one strategy is to push clean samples close to their corresponding adversarial samples in the feature space such that the prediction for clean and adversarial samples are coherent and thus the model is less vulnerable to adversarial attacks. 
    % This strategy is equivalent to 
    We achieve this by adding a positive link for each $\vx_i$ to its corresponding adversarial sample $\vx_i^\prime$ in ATG. We denote this set of positive links as $\cE_{\text{ca}}^+$. Each edge connects a clean sample $\vx_i$ and its corresponding adversarial sample $\vx_i^\prime$ with a link weight $\eta_1$. 
    
    \textbf{2) $\cE_{\text{intra}}^+$}: To boost intra-class feature similarity, one intuitive way is to enforce samples from the same class to be close in the feature space. Therefore, for a node $\vx_i$ with the label $y_i$, we connect it with all samples (including both clean and adversarial sample) from the same class $y_i$ excluding its adversarial sample. We denote this set of ``intra-class'' positive links, which is $\cE^+ \backslash \cE_\text{ca}^+$ , as $\cE_\text{intra}^+$ with a link weight $\eta_2$. 
    \item \textbf{Negative links} $\cE^-$. To increase inter-class feature variance, we construct negative links between two nodes from different classes in ATG. We assign each negative link with a weight $\eta_3$.
    %We denote this type link as $\cE^- = \{ (\vx_i, \vx_j), (\vx_i, \vx_j') , (\vx_i', \vx_j') | y_i \neq y_j \text{ and } i \neq j \}_{i,j=1}^{n}$. 
\end{itemize}

\begin{figure*}[t]
\centering
\subfloat[$\cE_{\text{ca}}^+$]{
\includegraphics[width = 0.3\textwidth]{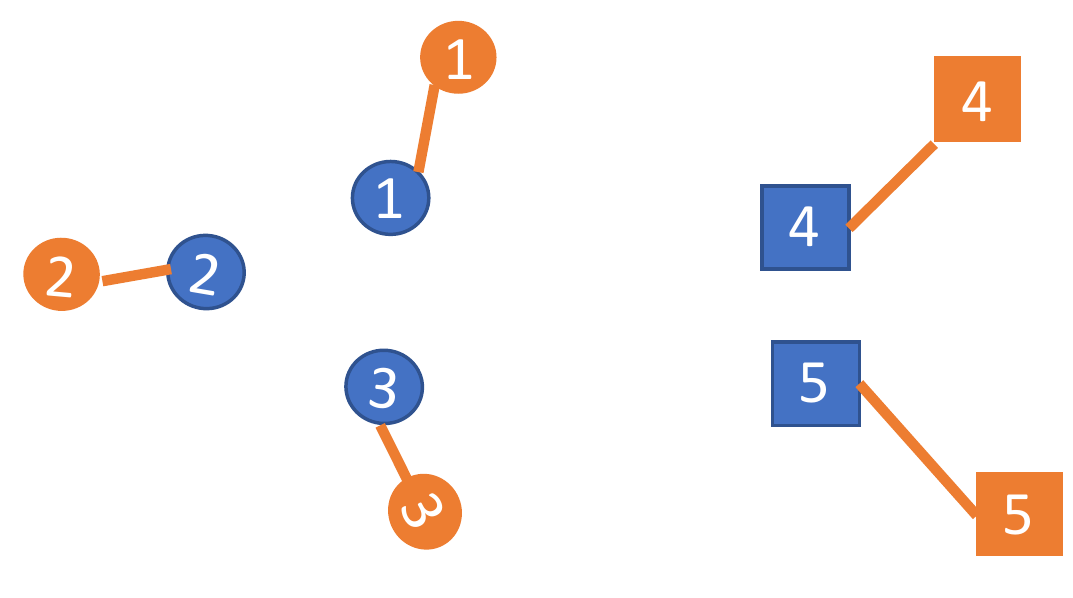}\label{fig:plca}} 
\hspace{0.22in}
\subfloat[$\cE_{\text{intra}}^+$]{\includegraphics[width = 0.3\textwidth]{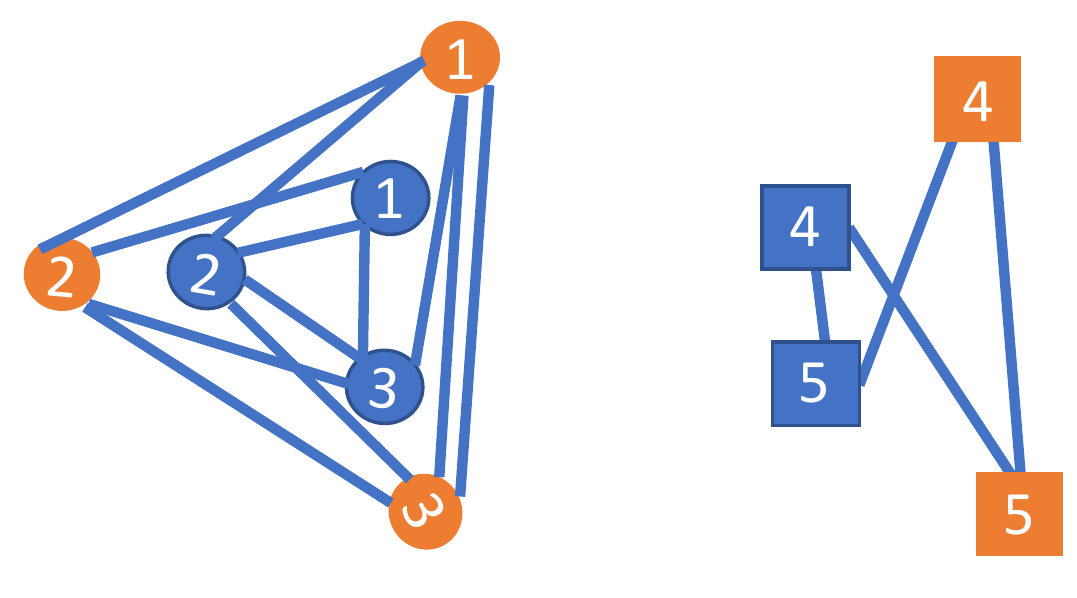}\label{fig:plintra}}
\hspace{0.22in}
\subfloat[$\cE_{\text{inter}}^-$]{\includegraphics[width = 0.3\textwidth]{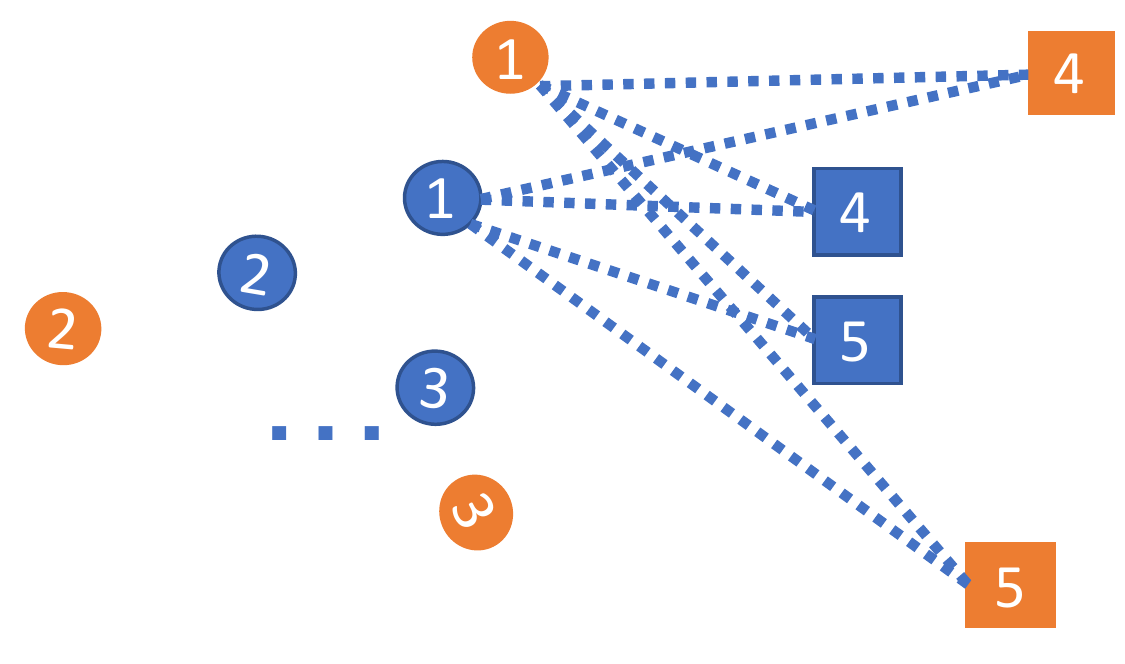}\label{fig:ngitra}}

\caption{An illustration of links in ATG. We use blue and orange color to denote clean and adversarial samples respectively, and circles and squares to indicate classes $a$ and $b$ separately. The orange solid lines between adversarial samples are positive links in $\cE_{\text{ca}}^{+}$, the solid blue lines are positive links in $\cE_{\text{intra}}^{+}$, and the dotted lines are negative links in $\cE^{-}$. Note that for clarity, we only show the negative links between the sample $1$ from the class $a$ to samples from the class $b$.} 
\label{fig: ATG}
\end{figure*}

In ATG, we have two types of positive links which allows us to enforce different levels of similarity by controlling their link weights $\eta_1$ and $\eta_2$. Meanwhile ATG is a complete graph which indicates that any pair of nodes in ATG is connected via either a positive link or a negative link. In the remaining of the paper, we use $\cE_{\text{ca}}^+ (i)$, $\cE_{\text{intra}}^+(i)$ and $\cE^-(i)$ to denote the three types of links for a node $\vx_i$ and  define $\cE (i)$ as $\cE (i) = \cE_{\text{ca}}^+ (i) \cup \cE_{\text{intra}}^+(i) \cup \cE^-(i)$ to indicate all links connecting with $\vx_i$. Suppose that we have a dataset with $5$ samples $\{1,2,3,4,5\}$ from $2$ classes $\{a,b\}$ where $\{1,2,3\} \in a$ and $\{4,5\} \in b$. We use $\{1', 2', 3', 4', 5'\}$ to denote their corresponding adversarial samples. An illustration of links in ATG for this toy training dataset is demonstrated in Figure~\ref{fig: ATG}, where Figure~\ref{fig:plca} denotes positive links between clean samples and their corresponding adversarial samples, Figure~\ref{fig:plintra} indicates positive links from the same class excluding $\cE_{\text{ca}}^+$ and Figure~\ref{fig:ngitra} shows negative links. Note that for clarity, Figure~\ref{fig:ngitra} only shows negative links between the sample ``1" from class $a$ and samples from class $b$.

\subsection{Capturing Link Information in ATG}
With the constructed ATG, the aforementioned three goals can be unified and simultaneously achieved by capturing link information in ATG. In general, ATG provides the guidance on the desired similarity of different samples around a center node $\vx_i$. Specifically, we aim to maximize the representation similarity of samples connected by positive links and minimize the similarity of samples connected by negative links. Thus, we desire to learn a model with a representation function $h(\cdot)$ that has large $\mathcal{L}_\text{FS} (h, ATG, \vx_i)$, which is defined as:
\begin{align*}
\begin{split}
  \mathcal{L}_\text{FS}(h, ATG, \vx_i) = & \;  \sum\limits_{(\vx_i, \vx_i')  \in \cE_\text{ca}^+(i)} s(h(\vx_i), h(\vx_i')) \\
   + & \; \sum\limits_{(\vx_i, \vx_j)  \in \cE_\text{intra}^+(i)} s( h(\vx_i), h(\vx_j)) \\
   - & \;  \sum\limits_{(\vx_i, \vx_j)  \in \cE^-(i)} s( h(\vx_i), h(\vx_j)),
\end{split}
%\label{l_atg}
\end{align*}

\noindent where $s(\cdot,\cdot,\cdot)$ is a similarity function. By maximizing $\mathcal{L}_\text{FS} (h, ATG, \vx_i)$ on the whole node set $\mathcal{V}$, we can achieve these goals on all samples. 
%\xr{here function $s$ can model the probability of the existence of any link, including positive and negative links.}
In this work, we define the function $s(\cdot, \cdot,\cdot)$ in $\mathcal{L}_\text{FS} (h, ATG, \vx_i)$ to measure the first-order proximity of a positive link in ATG. In particular, it is mathematically stated as the probability of the existence of a positive link among all links connected with the node $\vx_i$, and modeled with a softmax function. Using a positive link $(\vx_i, \vx_i') \in \cE_\text{ca}^+(i)$ as an illustrative example, $s(h(\vx_i), h(\vx_i'))$ is defined as
\begin{align}
    \begin{split}
        &s(h(\vx_i), h(\vx_i')) =  \frac{ \text{exp}( h(\vx_i)^\text{T} h(\vx_i'))}{S},\\
    \end{split}
    \label{link_ca}
    \end{align}
where 
\begin{align*}
% \begin{split}
  S& = \sum\limits_{(\vx_i, \vx_i')  \in \cE_\text{ca}^+(i)} \!  \text{exp}( h(\vx_i)^\text{T} h(\vx_i')) \\
  &+ \! \sum\limits_{(\vx_i, \vx_j)  \in \cE_\text{intra}^+(i)} \!  \text{exp}( h(\vx_i)^\text{T} h(\vx_j))\\  
  &+ \! \sum\limits_{(\vx_i, \vx_j)  \in \cE^-(i)} \! \text{exp}( h(\vx_i)^\text{T} h(\vx_j)).
% \end{split} \nonumber
\end{align*} 

Given that the representation $h(\vx_{i})$ is normalized, $h(\vx_{i})^{\text{T}} h(\vx_{i}')$ essentially denotes the cosine similarity of $h(\vx_{i})$ and $h(\vx_{i}')$. Thus maximizing $s( h(\vx_{i}), h(\vx_{i}'))$ can naturally enforce large similarity between $h(\vx_{i})$ and $h(\vx_{i}')$. $S$ is a normalization term over all links connecting to $\vx_{i}$. With the definition of $s(\cdot, \cdot)$, $\mathcal{L}_\text{FS}(h, ATG, \vx_{i})$ can be rewritten as 
% \begin{align}
% \begin{split}
%   \mathcal{L}_\text{FS}(h, ATG) = & \; \eta_1 \sum\limits_{(\vx_i, \vx_i')  \in \cE_\text{ca}^+} \frac{\text{exp}(  h(\vx_i)^\text{T} h(\vx_i'))}{\sum\limits_{(\vx_i, \vx_k) \in 
%   \cE} \text{exp}( h(\vx_k)^\text{T} h(\vx_i))} \\
%   + & \; \eta_2 \sum\limits_{(\vx_i, \vx_j)  \in \cE_\text{intra}^+} \frac{\text{exp}(  h(\vx_i)^\text{T} h(\vx_j))}{\sum\limits_{(\vx_i, \vx_k) \in \cE} \text{exp}( h(\vx_k)^\text{T} h(\vx_i))} \\
%   - & \; \eta_3 \sum\limits_{(\vx_i, \vx_j)  \in \cE^-} \frac{\text{exp}(  h(\vx_i)^\text{T} h(\vx_i'))}{\sum\limits_{(\vx_i, \vx_k) \in \cE} \text{exp}( h(\vx_k)^\text{T} h(\vx_i))}. 
% \end{split}
% \label{l_atg_2}
% \end{align}

\vspace{-1em}
\begin{align}
\begin{split}
  \mathcal{L}_\text{FS}(h, ATG, \vx_i)  = & \frac{1}{S}  \left(  \sum\limits_{(\vx_i, \vx_i')  \in \cE_\text{ca}^+(i)}   \text{exp}(h(\vx_i)^\text{T} h(\vx_i')) \right.\\
  & +  \sum\limits_{(\vx_i, \vx_j)  \in \cE_\text{intra}^+(i)}   \text{exp}(h(\vx_i)^\text{T} h(\vx_j)) \\
  &\left. -  \sum\limits_{(\vx_i, \vx_j)  \in \cE^-(i)}   \text{exp}( h(\vx_i)^\text{T} h(\vx_i'))  \right).
\end{split}
\label{l_atg_2}
\end{align}

Note that the normalization denominator $S$ is the summation of the numerators of the three terms in Eq.~(\ref{l_atg_2}). Thus, one advantage of defining $s(\cdot, \cdot,\cdot)$ as Eq.~(\ref{link_ca}) is -- maximizing the first two terms for positive links in Eq.~(\ref{l_atg_2}) will lead to minimizing the third term for negative links automatically. 
%\xr{Can we formally explain why? For instance, $\argmax \frac{A+B-C}{A+B+C} = \argmax \frac{A+B-C}{A+B+C}+1 =\argmax \frac{2(A+B)}{A+B+C} = \argmax \frac{A+B}{A+B+C}$} 
Therefore, we can further simplify  $\mathcal{L}_\text{FS}(h, ATG, \vx_i)$  in Eq.~(\ref{l_atg_2}) as the mean of log-likelihood of all positive links connected by $\vx_i$ and add different weights to different types of links. Since different nodes in ATG can have different numbers of positive links, we add a normalization term $\frac{1}{|\cE^+(i)|}$ in $\mathcal{L}_\text{FS} (h, ATG, \vx_i)$. As a result, the feature separability loss is defined as follows:

\begin{align}
\begin{split}
   \mathcal{L}_\text{FS} (h, ATG, \vx_i) &= \\ 
  \frac{1}{|\cE^+(i)|} & \left( \eta_1 \cdot \sum\limits_{(\vx_i, \vx_i')  \in \cE_\text{ca}^+(i)} \text{log } \frac{ \text{exp}(  h(\vx_i)^\text{T} h(\vx_i'))}{S} \right. \\
   +  &\eta_2 \cdot \left.   \sum\limits_{(\vx_i, \vx_j)  \in \cE_\text{intra}^+(i)} \text{log } \frac{\text{exp}( h(\vx_i)^\text{T} h(\vx_j))}{S} \right), \\
\end{split}
\label{l_atg_3_old}
\end{align}

\subsection{The ATFS Algorithm}

The final objective function for Adversarial Training with Feature Separability (ATFS) is a combination of the cross-entropy loss and the objective provided by ATG: 
\begin{align}
% \begin{split}
%   \min_{f_{\theta}} \sum\limits_{x_i}  \alpha\cdot \mathcal{L}_{\text{CE}}(f(\vx'_i), \vy_i) - \beta\cdot \mathcal{L}_{\text{FS}}(F(\cdot), ATG, \vx_i).
   \min_{f_{\theta}} \sum\limits_{x_i\in\cD}  \alpha\cdot \mathcal{L}_{\text{adv}}(f(\vx'_i), y_i) - 
   \sum\limits_{x_i\in\cV} \beta\cdot \mathcal{L}_{\text{FS}}(F(\cdot), ATG, \vx_i),
% \end{split}
\label{objective}
\end{align}

\noindent where $F(\cdot)$ denotes the features extracted by the model, which is the output of the second last layer of the model. The first term in Eq.~(\ref{objective}) is the cross-entropy loss, which minimizes the error risk for the adversarial samples towards the correct labels. Here we have the flexibility to calculate $\mathcal{L}_{\text{adv}}$ based on different adversarial training algorithm. The second term is the Feature Separability Loss which facilitates the model to learn separable feature representations. Two parameters $\alpha$ and $\beta$ are predefined to balance the contributions from these two terms. In this work, we adopt SGD to optimize the proposed objective function. The detailed training algorithm for ATFS can be found in Appendix A. 

\section{Experiment}
\label{sec:experiment}

% \iffalse
% \jt{I do not think we should start the section as it. Typically we can say:  In this section, we conduct experiments to validate the effectiveness of the proposed framework. We first introduce the experimental settings, then evaluate the robustness of the proposed framework compared to representative baselines,  next understand xxx  and finally present case studies..... 
% }
% \fi

%In this section, we conduct comprehensive experiments to evaluate the performance of proposed ATFS on benchmark datasets with ResNet18 and WideResNet-34-10 model. Then, We also show how the robust performance changes along with the change of weight for the FS loss term to further demonstrate the effectiveness of ATG in the ablation study. Besides, we try to understand how the proposed ATG help to learn better feature representation by feature visualization presented in case study.

In this section, we conduct comprehensive experiments to evaluate the effectiveness of the proposed ATFS framework. We first introduce the experimental settings and then compare ATFS with representative baselines in terms of clean and robust performance. Next we conduct ablation study to show the impact of model components on ATFS and finally further probe the advantages of ATFS. 

%. Then, We also show how the robust performance changes along with the change of weight for the FS loss term to further demonstrate the effectiveness of ATG in ablation study. Besides, we try to understand how the proposed ATG help to learn better feature representation by feature visualisation presented in case study.
\setlength{\tabcolsep}{8pt}{
\begin{table*}[t]
\caption{Performance Comparison on CIFAR10 with ResNet18.}
\label{tab:cifar10-resnet18}
\centering
\begin{tabular}{@{}l|c|c|c|c|c@{}}
\hline
\hline
& Clean & FGSM & PGD-20(8/255) &  $\text{CW}_{\inf}$(8/255) & AA(8/255)   \\
\hline
%Clean Model     &   -    &  0     &    0    &     0    &  0  \\
AT                      & 84.27 & 62.66±0.07 & 49.44±0.03   & 48.18±0.00 & 46.30 \\
TRADES ($1/\lambda=5.0$)   & 81.64   & 61.78±0.03  & 50.55±0.03      & 48.41±0.03 & 47.95 \\
%MLA &   &   &     &   & \\
RoCL+AT (finetune)      &  80.26 & 58.14±0.00   & 40.77±0.00    &  40.12±0.00 & 40.12 \\
SS-OOD & 80.58 & 61.21±0.03 & 49.64±0.03&   48.54±0.00 & 46.67\\ 
MART & 83.07  & 65.43±0.07  & 53.25±0.03  &  49.46±0.00   & 47.93\\
%GAIRAT &  & &  & &\\

%AWP & 80.80 & 64.30±0.05 & 54.53±0.05 &  51.62±0.06   &  \\
\hline
%clean+CL($\alpha/\beta= 100$,$\eta$ = 1)     & \textbf{85.89} & 50.93±0.004  & 15.89   &    48.97      & 12.22    \\
%ATFS(clean)   & 84.58 & 65.21±0.05 & 52.49±0.06    &   &   51.49±0.04  \\
%ATFS(clean) ($\beta/\alpha= 100$, $\eta = 3$)   & \textbf{85.64} &   \textbf{65.70±0.03} & 51.96±0.03      &  51.02±0.04 &   51.04±0.03\\
AT+ATFS     & 83.74 & 64.64±0.03 & 51.99±0.00    & 53.02±0.00 & 48.03 \\
%ATFS(adv) ($\beta/\alpha = 100$, $\eta = 3$) & 83.80 & 64.91± 0.07 & 53.34±0.06   & 52.47±0.00 &   52.44±0.03 \\ 
%AWP + ATFS(clean) & 85.54 & 67.64±0.00 &  55.76±0.03   &  &  47.19+0.05\\
TRADES + ATFS & 81.88 &  63.26±0.03 & 52.24±0.00 &  49.72±0.05  & 48.66 \\

MART + ATFS & 82.79 & 64.00±0.00  & 54.50±0.00 & 50.12±0.00 & 48.39\\

%AWP + ATFS& 86.11  &  67.28±0.06 & 54.74±0.00 &  & 47.44+0.06 \\

\hline
\hline
\end{tabular}
\end{table*}
}

\begin{table*}[h]
\caption{Performance Comparison on SVHN with ResNet18.}
\label{tab:svhn-resnet18}
\centering
\begin{tabular}{@{}l|c|c|c|c|c@{}}
\hline
\hline
       & Clean & FGSM & PGD-20(8/255)  &   $\text{CW}_{\inf}$(8/255)  & AA(8/255)   \\
\hline
AT     & 91.49 &  63.38±0.00 &  52.71±0.00  &  48.64±0.00   &  44.07      \\
TRADES ($1/\lambda=2.0$)   & 90.84 &  71.09±0.03   & 54.79±0.03  &   51.24±0.00 & 47.36\\
TRADES ($1/\lambda=5.0$)   &  89.82 &   71.97±0.00   &  56.83±0.03  & 52.83±0.00  &  48.14\\
%MLA &  &      &   &  & \\
SS-OOD & 90.62 & 69.28±0.00 & 51.45±0.00 &  47.75±0.03  & 44.76 \\ 
MART & 92.02 & 74.41±0.03 & 57.76±0.03 &  51.18±0.03 & 47.20 \\
%GAIRAT &  & &  & &\\

%AWP & 90.97 & 69.08±0.03 & 55.87±0.00   & 43.02±0.03 &  41.26 \\
\hline
%clean+CL($\alpha/\beta= 100$,$\eta$ = 1)     & \textbf{85.89} & 50.93±0.004  & 15.89   &    48.97      & 12.22    \\

AT + ATFS  & 89.95  & 72.79±0.07  & 58.82±0.00  &    52.77±0.04 & 45.89\\
%ATFS($\alpha/\beta$ = 100, $\eta$ =3)  & 79.49  & 59.95±0.000   &  49.73±0.001 &  48.46  & 46.24±0.000 \\
TRADES + ATFS &    90.30     &    73.55±0.02   &   58.86±0.02   &  54.35±0.04   &  49.93 \\
MART + ATFS &    91.79     &   74.17±0.01   & 57.25±0.02       & 51.06±0.00&   48.40     \\
%AWP + ATFS &   88.22       &  69.22±0.05      &  56.60±0.00     &    53.89     &   51.38±0.00\\
\hline
\hline
\end{tabular}
\end{table*}

\subsection{Experimental Settings}
%CIFAR100~\cite{krizhevsky2009learning}
In this section, we introduce the experimental setting including the details of the training and test phase.

\textbf{Training Setup} To demonstrate the effectiveness of the proposed framework, we conduct experiments on two model architectures, including ResNet18~\cite{he2016deep} and WRN-34~\cite{zagoruyko2016wide}. We test the performance on two benchmark datasets, including CIFAR10~\cite{krizhevsky2009learning} and SVHN~\cite{netzer2011reading}. 
For CIFAR10, we use 49,000 images for training and the rest 1,000 images for validation. For SVHN, we use 72,257 images for training and 1,000 for validation. We train each model for a maximum of 120 epochs. While, recent study~\cite{rice2020overfitting} suggests using early stopping because adversarial training suffers from serious overfitting issues. Therefore, we split the training set into a training set and a validation set, and use the validation set to select the best model for all experiments and baselines.  All the models are trained using SGD with momentum 0.9, weight decay $2 \times 10^{-4}$ and an initial learning rate of 0.1, which is divided by 10 at the 75-th and
90-th epoch. For the parameter selection, we set the weight $\beta$ for the $\mathcal{L}_adv$ as 1 and select the weight $\alpha$ for FS loss from $\{0.01, 0.05, 0.1, 0.2, 0.5\}$. Meanwhile, for the parameters in the ATG, we set $\eta_2=\eta_3=1$ and keep $\eta_1$ flexible to ease the tuning process for our model. The adversarial samples used in training are calculated by PGD-$10$, with a perturbation budget $\epsilon=8/255$, step size $\gamma = 2/255$ and step $10$. 

\textbf{Robust Evaluation} For robustness evaluation, we report robust accuracy under $l_\infty$-norm 8/255 attacks generated by various attacking algorithms including FGSM attack~\cite{goodfellow2014explaining}, PGD attack~\cite{madry2017towards}, CW Attack optimized with PGD ~\cite{carlini2017towards} and Autoattack (AA)~\cite{croce2020reliable}. Note that AA is one of the strongest attack algorithms that ensembles three white-box attacks (APGD-CE, APGD-DLR, FAB) and one black-box attack (Square Attack). These attack methods are based on the implementation by DeepRobust~\cite{li2020deeprobust} and Autoattack~\cite{croce2020reliable}. 

\subsection{Robust Performance Comparison}

To assess the robustness achieved by ATFS, we compare the performance of ATFS with 
% numerous representative baselines including
four adversarial training methods, including standard adversarial training (AT)~\cite{madry2017towards},  TRADES~\cite{zhang2019theoretically} and MART ~\cite{wang2019improving}. Two self-supervised related adversarial training methods SS-OOD (AT + Auxiliary Rotation)~\cite{hendrycks2019using} and  RoCL~\cite{Kim2020}.
% Moreover, AWP~\cite{wu2020adversarial} is one recent proposed method which not only train the model by adversarial training but also adds designed perturbation on the model parameters to increase robust performance. 
% It can be incorporated into different adversarial training methods. Thus, we combine AWP with ATFS to further improve generalization. 

{\bf Performance Comparison on ResNet18}
Table~\ref{tab:cifar10-resnet18} shows the performance comparison on the CIFAR10 dataset and we can make the following observations. First, ATFS outperforms two classic adversarial training methods (AT and TRADES). For instance, in terms of Autoattack, the robust accuracy of ATFS improves over AT and TRADES ($1/\lambda = 5.0$) by $1.7\%$ and $0.71\%$ respectively, while maintaining similar or better clean accuracy. Note that TRADES also explores the relations among clean and adversarial samples, while ATFS explores not only the relationship between clean and adversarial samples but also the class wise relations via ATG. This comparison indicates that capturing the relations of inter and intra class samples can boost both the clean and robust performance. 
ATFS also significantly outperforms two self-supervised related adversarial training methods. In terms of PGD-20 attack, the robust accuracy of AT+ATFS improves over RoCL (AT finetune) and SS-OOD by $10.22\%$ and $2.35\%$, respectively. 
When advanced adversarial training algorithms like TRADES and MART are combined with ATFS, the performance can be further improved. In principle, those methods improve adversarial training in different ways from ATFS and their contributions could be complementary. Therefore, a combination of them has the potential to further boost performance. %We provide such an example through the combination of ATFS and AWP. The objective is defined as:

The performance comparison on the SVHN dataset is shown in Table~\ref{tab:svhn-resnet18} and we can make similar observations as those on the CIFAR10 dataset:
AT+ATFS achieves better clean and robust performance than AT and TRADES in most of the cases. It also outperforms the self-supervised baseline SS-OOD by $7.41\%$ under the PGD-20 attack. To combine with MART and TRADES, the performance is further improved. The combination of AT, TRADES, and MART is improved by $1.82\%$, $1.79\%$ and $1.2\%$ compared to the original algorithm.

% illustrates the robust performance on the SVHN dataset. 
% ATFS can achieve comparable or better performance compared to the SOTA methods against all the attack methods. 

{\bf Performance on WRN-34-10.} To test the performance of ATFS with larger model capacity, we train it on WideResNet-34~\cite{zagoruyko2016wide} and we have similar observations with ResNet18: Adversarial training algorithms can achieve better performance with feature separability. The results can be found at Table~\ref{tab: CIFAR10_WRN34} in Appendix B. 
%From the table, we can make the following observations:
% Compared to AT and TRADES, ATFS can achieve both better clean and robust performance. This observation indicates that ATFS can increase robustness without sacrificing clean performance compared to TRADES. While, compared to MART, the performance improvement is less significant. significantly outperform SS-OOD. In this setting, we do not show the experiment of RoCL since this method is trained in a totally unlabeled manner, which requires much longer training epochs to converge, and thus the total running time is not affordable.
    
% ATFS(clean) and ATFS(adv) achieve better or comparable performance under different attacks compared with with the state-of-the-art methods MART. 
    
% For the combination method with AWP, the performance is improved compares to ATFS(adv) or ATFS(clean). Specifically, ATFS(adv) + AWP achieves the best robust performance among all baselines under Autoattack.

\subsection{Ablation Study}
\label{sec:ablation}
In this subsection, we aim to provide analysis on why ATFS can help both clean and robust performance. In particular, we study how the model components affect the performance of ATFS.

\subsubsection{Effect of the FS Loss}
\begin{figure}[!h]
\centering
\subfloat[FS Loss Weight $\beta$]{\label{cifar10_weight}\includegraphics[width = 0.23\textwidth]{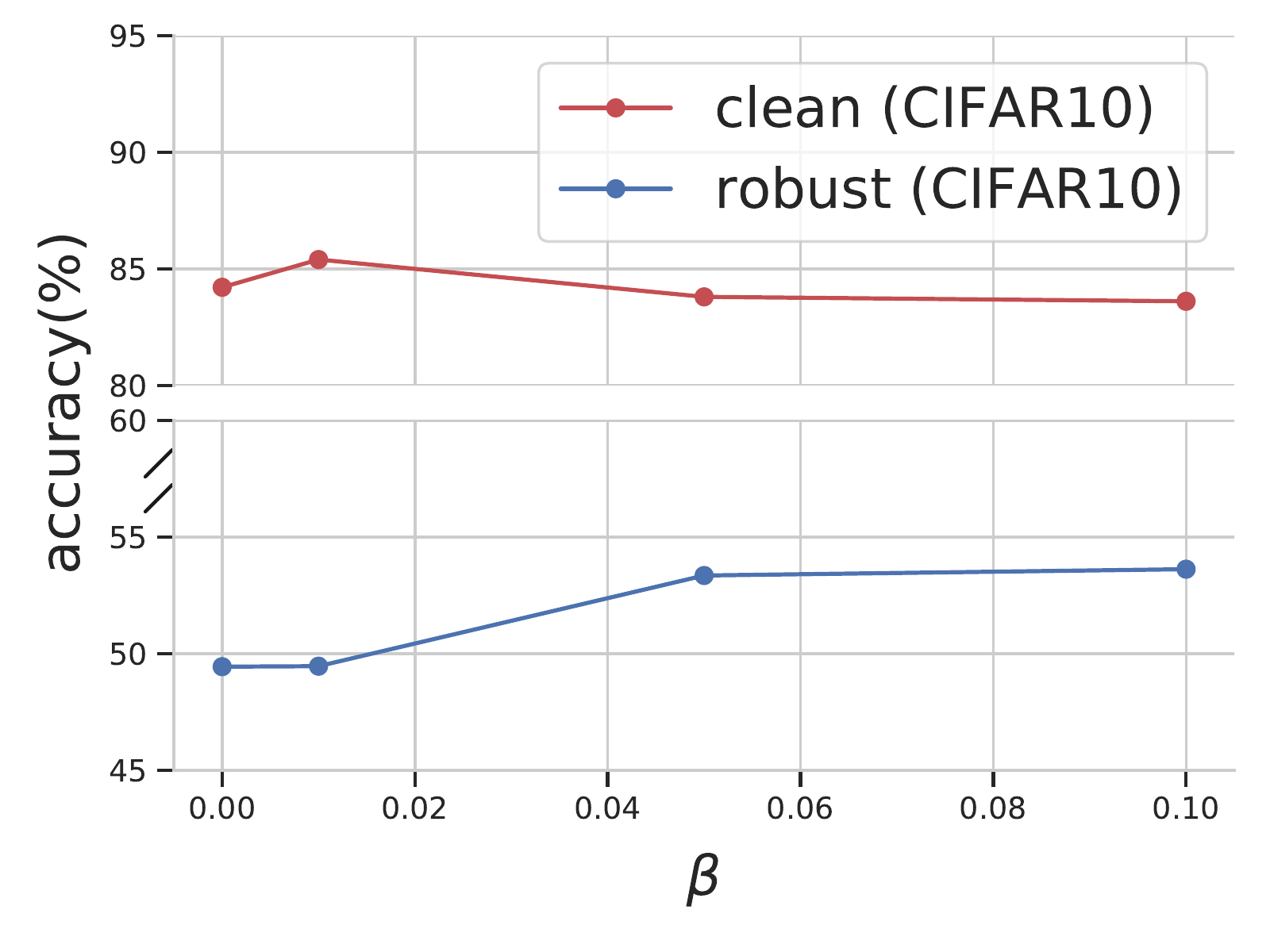}}
\subfloat[Linkweight $\eta$]{\label{cifar10_linkweight}\includegraphics[width = 0.23\textwidth]{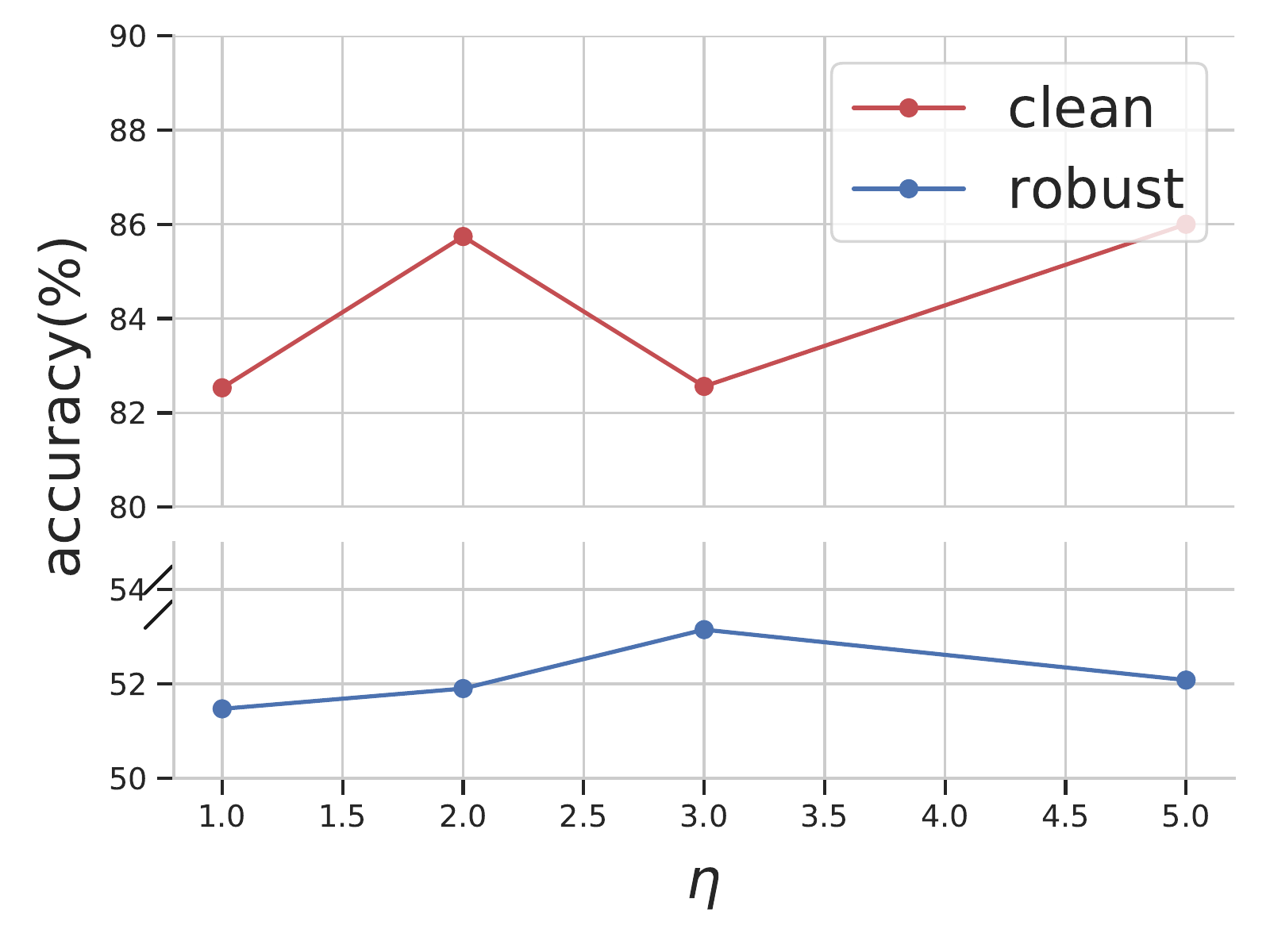}}
\caption{The impact of the FS loss and adversarial link weight}
\label{fig:ablation_weight}
\end{figure}

The training objective defined in Eq.~(\ref{objective}) is the combination of the adversarial loss $\mathcal{L}_{adv}$ and the proposed FS loss $\mathcal{L}_{FS}$. To study how $\mathcal{L}_{FS}$ impacts the performance, we conduct experiments to see how the performance changes with the weight for $\mathcal{L}_{FS}$. We fix the weight parameter $\alpha = 1$ and changes $\beta$ from {0.05, 0.1, 0.5, 1.0}. The results on CIFAR10 is shown in Figure~\ref{cifar10_weight}. 

% \begin{figure}[!h]
% \centering
% \subfloat[CIFAR10]{\label{cifar10_weight}\includegraphics[width = 0.24\textwidth]{pic/ablation_weight_cifar10.pdf}}
% \subfloat[SVHN]{\label{svhn_weight}\includegraphics[width = 0.24\textwidth]{pic/ablation_weight_svhn.pdf}} 
% \caption{The impact of the FS loss.}
% \label{fig:ablation_weight}
% \end{figure}

In this figure, the red lines denote the clean performance and blue lines indicate the robust performance. We can observe that with the increase of the weight for $\mathcal{L}_{FS}$, the robust performance increases $4.2\%$ compared to standard adversarial training ($\beta = 0$) on CIFAR10 while we do not observe an obvious decrease for clean accuracy. 
This is consistent with the result in SVHN dataset. 
%For SVHN dataset,  with the increase of the weight for $\cL_{\text{FS}}$, the clean and robust accuracy first increase and then start to drop after some values. 
%This study validates that the $\mathcal{L}_{FS}$ can contribute to better clean and robust performance. Thus it further demonstrates the effectiveness of exploring various types of relations via ATG. On the other hand, continuing to increase the weight for the $\cL_{\text{FS}}$ will not lead to a continuous increasing of the robust performance, which indicates the necessity of the supervised cross-entropy loss. 
% to provide the label information. \xr{L_FS also uses label information to construct ATG}

\subsubsection{Effect of Adversarial Link Weight in ATG}

% \begin{figure}[!h]
% \centering
% \subfloat[AT+ATFS]{\label{cifar10_weight}\includegraphics[width = 0.24\textwidth]{pic/ablation_linkweight.pdf}}
% \subfloat[TRADES+ATFS]{\label{svhn_weight}\includegraphics[width = 0.24\textwidth]{pic/ablation_weight_svhn.pdf}} 
% \caption{The impact of the Adversarial Link Weight}
% \label{fig:ablation_weight}
% \end{figure}

% \begin{figure}[!h]
% \centering
% \subfloat[AT+ATFS]{\label{cifar10_weight}\includegraphics[width = 0.24\textwidth]{pic/ablation_linkweight.pdf}}
% \caption{The impact of the Adversarial Link Weight}
% \label{fig:ablation_weight}
% \end{figure}

In ATG, we have two types of positive links, $\cE_{\text{ca}}^+$ and $\cE_{\text{intra}}^+$. Those two types of links are assigned with weights $\eta_1$ and $\eta_2$, separately. We show how the robust performance changes along with the $\eta_1$ changes in Figure~\ref{cifar10_linkweight}. We can see that when the link weight between clean samples and adversarial samples grow slightly larger, the robust performance will increase since the model focuses more on robust performance. While if $\eta_1$ increases even larger, it would degrade the performance since it will reduce the effect of feature separability. It shows that the feature separability loss will actually contribute to the final robust performance. 
%In Eq.~(\ref{objective}), there are two possible options for the cross-entropy loss. Since we minimize the feature difference between clean and adversarial samples in $\cL_{\text{FS}}$, $\cL_{\text{CE}}$ can either be calculated on clean samples (we denote this variant as ATFS(clean)) or adversarial samples (we denote this variant as ATFS(adv)) to guarantee that each sample is assigned to the correct label. To see whether both can achieve high clean and robust accuracy, we run the experiment for ATFS(adv) and ATFS(clean) with different $\eta_1 \in \{1,2,3\}$.
%Figure~\ref{CE_CIFAR10} and ~\ref{CE_svhn} show the results on CIFAR10 and SVHN, respectively. 

%In these two figures, the blue lines denote the robust performance and the red lines indicate the clean performance.
%Note that the robust performance of ATFS(adv) is slightly better, while ATFS(clean) achieves slightly better clean performance. 
%This suggests that ATFS(adv) (or ATFS(clean)) is a little bit bias towards robust (or clean) performance. Overall, both of them achieve good clean and robust accuracy, 
% It suggests that with the positive links $\cE_{\text{ca}}^+$ in ATG, both the adversarial samples or the clean samples are mapped into the correct labels, and correspondingly keep both good clean and robust performance. 
% It further indicates the effectiveness of capturing the positive links $\cE_{\text{ca}}$.
%which indicates the effectiveness of capturing the positive links $\cE_{\text{ca}}$ in ATG.

\subsection{Further Probing on Feature Space of ATFS}

In this subsection, we further probe the learned features to demonstrate the advantages of ATFS by visualizing these features and measuring the boundary thickness. 

\vspace{-1em}
\begin{figure}[!h]
\centering
\subfloat[$\beta = 0.01$]
{\includegraphics[width = 0.2\textwidth]{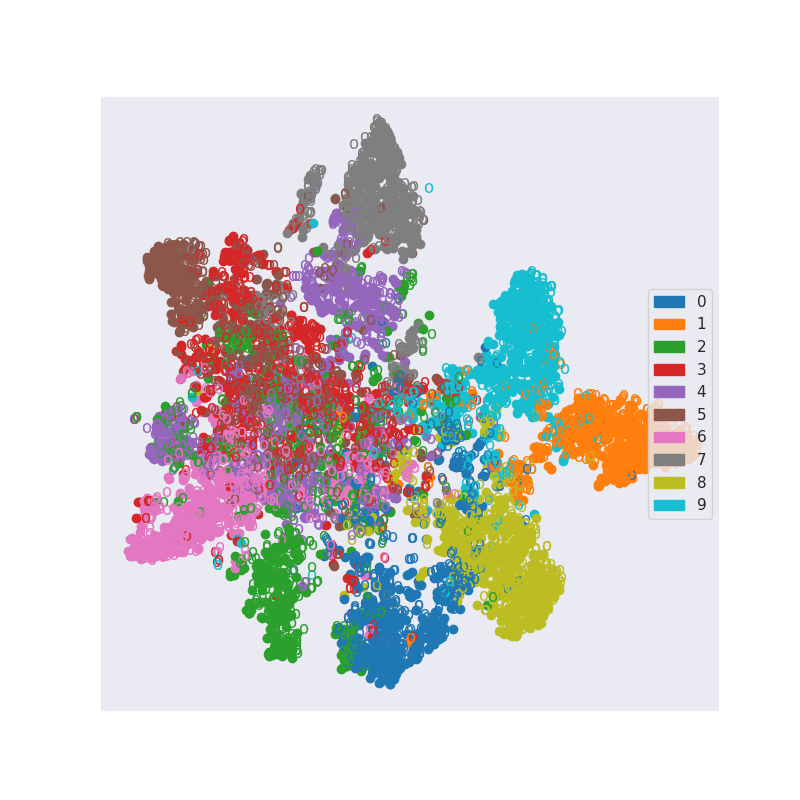}
\label{fig:pgd_visualize}} 
\subfloat[$\beta = 0.1$]
{\includegraphics[width = 0.2\textwidth]{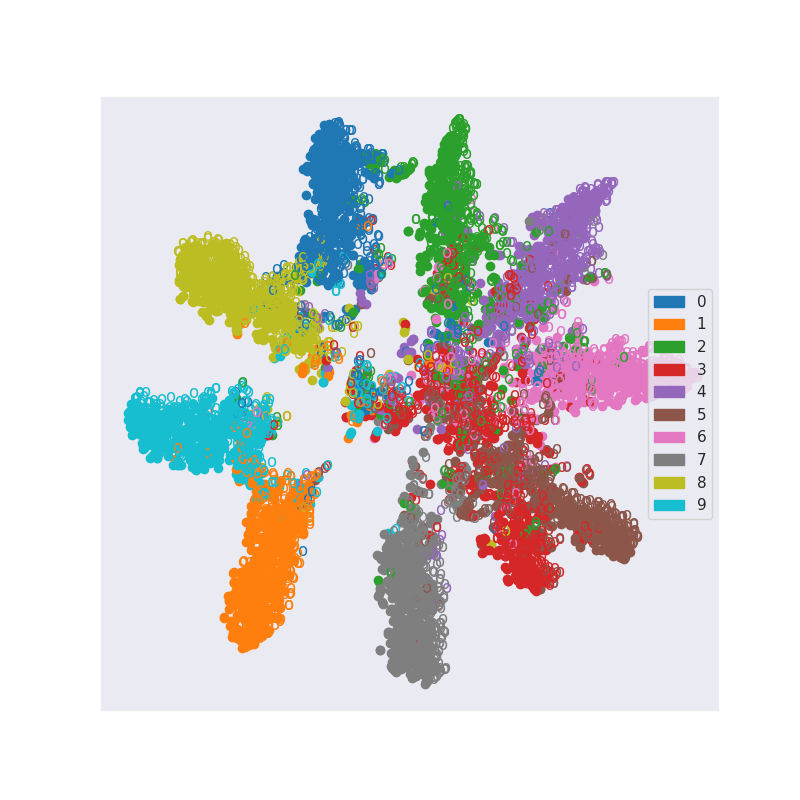}
\label{fig:atfs_visualize}}
\caption{Features learned by PGD adversarial training and ATFS where Figure~\ref{fig:pgd_visualize} shows the t-SNE feature visualization for the standard PGD adversarially trained model and Figure~\ref{fig:atfs_visualize} illustrates the feature visualization for the model trained by ATFS.}
\label{fig:similairty}
\end{figure}

\begin{figure}[t]
\centering
\subfloat[Feature correlation for AT]
 {\label{fig:correlation_AT}\includegraphics[width = 0.25\textwidth]{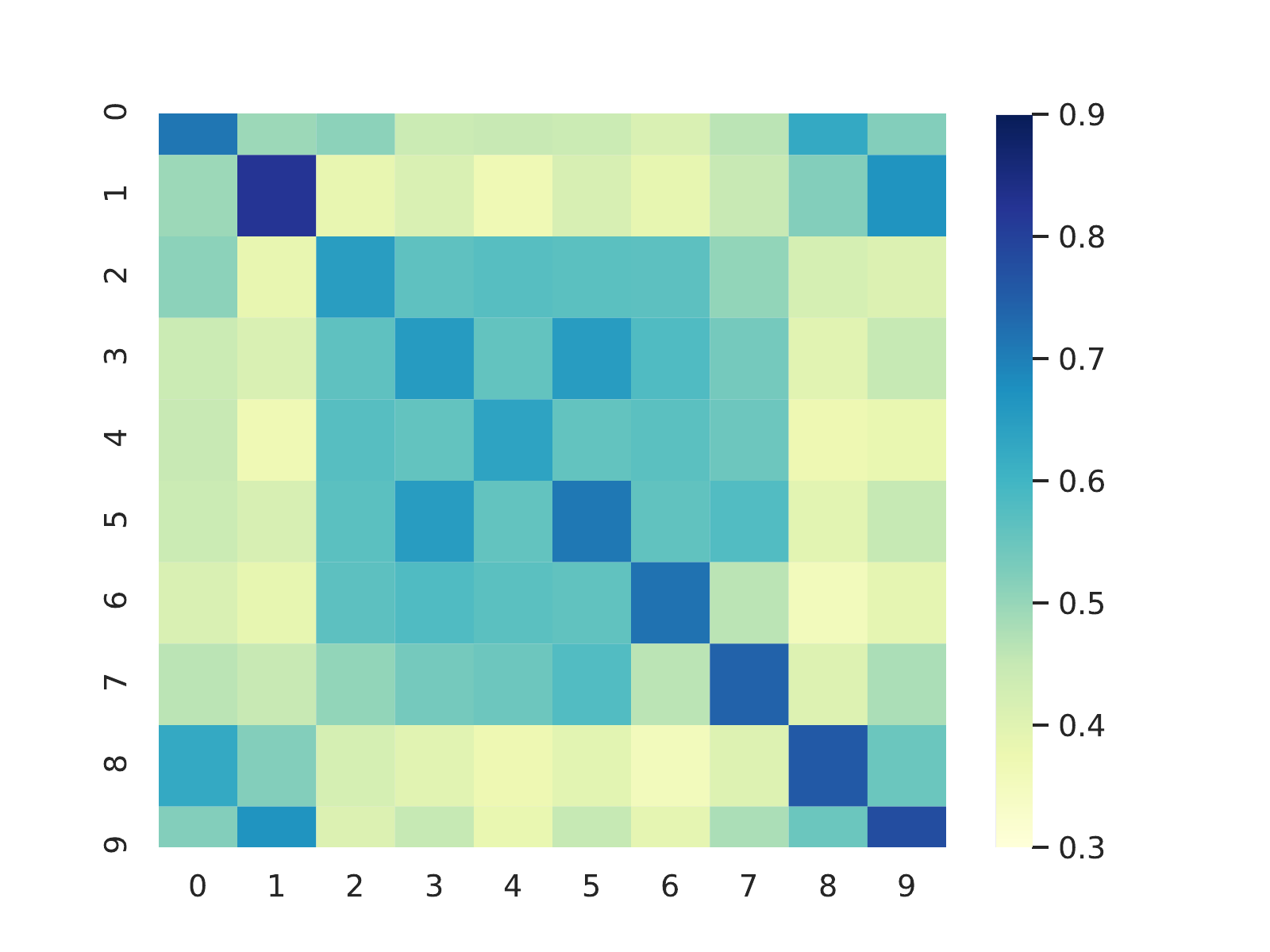}} 
 \subfloat[Feature correlation for ATFS]
 {\label{fig:correlation_ATFS}\includegraphics[width = 0.25\textwidth]{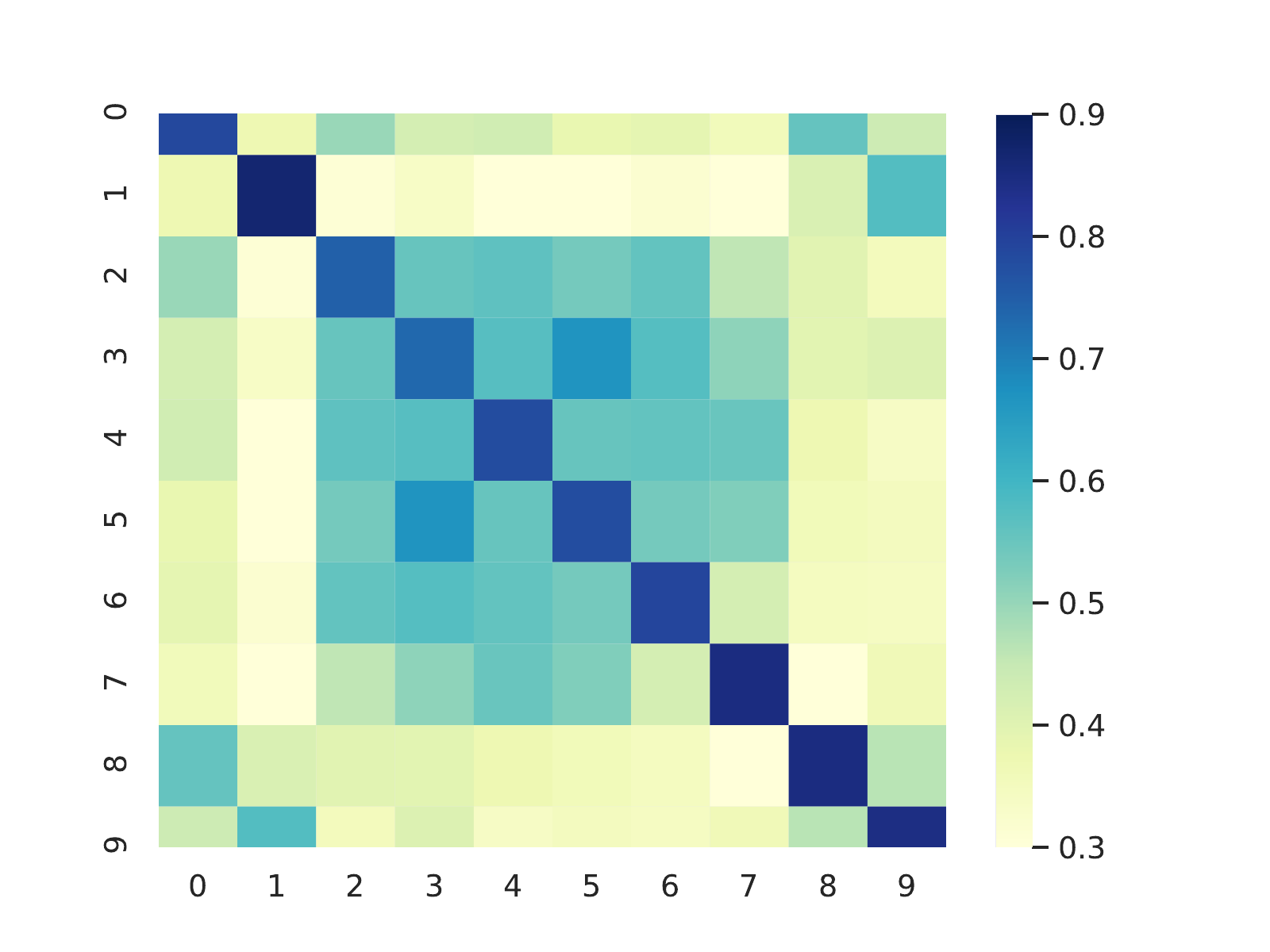}}
\caption{Feature similarity heatmap for the standard adversarial training and ATFS.} 
\label{fig:feature-similairty-heatmap}
\end{figure}
\vspace{-2em}

\subsubsection{Visualizing Learned Features} 
To give an illustration of how the FS loss term affects the feature space of the trained model, we show 1) t-SNE visualization of learned features  and 2) feature similarity heatmap among different classes.

% In Figure~\ref{fig:similairty},
Figure~\ref{fig:pgd_visualize} and Figure~\ref{fig:atfs_visualize} 
% the left subfigure is features learned by a standard AT model and the right one is features learned by ATFS. 
show the t-SNE feature visualization of standard adversarially trained model and the model trained by the ATFS framework when $\beta=0.01$ and $\beta=0.1$, respectively. For ATFS, representations of samples from different classes are well separated. Especially, clean samples and their corresponding adversarial samples in ATFS are closer to each other than AT. It demonstrates that, with the FS loss term, the objective can learn much better features and thus can explain why ATFS achieves good performance on both clean and robust samples. 

In addition to visualizing learned features, we also examine the average feature similarity for samples among different classes. As shown in Figure~\ref{fig:feature-similairty-heatmap}, the value in the $i$-th row and the $j$-th column denotes the average feature similarity of samples from classes $i$ and $j$, and darker color indicates higher similarity. Compared to standard adversarial training showed in Figure~\ref{fig:correlation_AT}, the value on the diagonal of the Figure~\ref{fig:correlation_ATFS} is larger, which suggests larger intra-class feature similarity for the model trained by the ATFS framework. On the other hand, the rest part of Figure~\ref{fig:correlation_ATFS} is lighter than Figure~\ref{fig:correlation_AT}, which indicates larger inter-class feature variance. To summarize, both the feature visualization and feature similarity verify that ATFS has achieved the aforementioned three goals and demonstrate why ATFS can achieve better performance.

\begin{table*}[t]
\centering
\caption{Measuring Boundary Thickness.}
\label{tab: thickness}
\begin{tabular}{c|c|c|c|c|c}
\hline
\hline
Method & 
AT & TRADES(2.0)& TRADES(5.0) & 
MART &
ATFS  \\
\hline
BT & 2.21 &   2.51    &  3.04  &   2.60   &    3.23                \\
\hline
\hline
\end{tabular}
\end{table*}
\subsubsection{Measuring Boundary Thickness} 

Boundary thickness (BT)~\cite{yang2020boundary} is a robustness measurement that intuitively indicates the cost of generating adversarial samples for a model. Basically, the boundary thickness of a classifier measures the expected distance to travel along with line segments between different classes across a decision boundary. It has been proven that a larger boundary thickness is necessary for robust models~\cite{yang2020boundary}. Therefore, we calculate the boundary thickness for different robust models to compare their robustness properties. The result is summarized in Table~\ref{tab: thickness}. We can observe that ATFS has the largest thickness value, This observation is consistent with the previous observations that ATFS can achieve 
the best robust performance.

\iffalse
\begin{figure}[h]
\centering
\begin{minipage}[SAT]{1\textwidth}
 \includegraphics[width = 0.4\textwidth]{pic/boundary.pdf}
\end{minipage}
\end{figure}
\fi

\section{Related Work}
\label{sec:related}
% This work focuses on the defense strategies based on adversarial training, so we mainly introduce related works about adversarial training in this section.

\textbf{Adversarial Attacks and Defenses} The existence of several successful attack methods, such as fast gradient sign method (FGSM)~\cite{goodfellow2014explaining}, CW attack~\cite{carlini2017towards} and Projected Gradient Descent (PDG) attack~\cite{madry2017towards}, demonstrated the vulnerability of deep neural networks to adversarial samples. To improve the model robustness against adversarial samples, many defensive methods have been proposed including input denoising~\cite{tramer2017ensemble, xu2020adversarial}, defensive distillation~\cite{papernot2016distillation}, gradient regularization~\cite{ross2018improving,tramer2017ensemble}, adversarial training~\cite{madry2017towards, goodfellow2014explaining} and certified defenses. Among them, adversarial training has been shown to be one of the most effective defenses~\cite{athalye2018obfuscated}. 
% The can effectively improve the model robustness will not be vulnerable to adaptive attacks~\cite{athalye2018obfuscated}. 
% Adversarial training can be formulated as solving a min-max optimization problem by feeding generated adversarial examples into the training process and minimizes model's error on the most dangerous examples. 
Based on adversarial training, a number of variants have been developed to further improve the performance, including TRADES~\cite{zhang2019theoretically},  MART~\cite{wang2019improving}, MMA training~\cite{ding2018mma}, FAT~\cite{zhang2020attacks}, GAIRAT~\cite{zhang2020geometry}.
Even though adversarial training and its variants are effective to improve model robustness, the adversarial training based methods still have several main drawbacks, such as the trade-off between clean accuracy and robust accuracy, which is first claimed in ~\cite{zhang2019theoretically}; bad clean performance generalization~\cite{schmidt2018adversarially} and robustness generalization~\cite{rice2020overfitting}, introducing fairness issue. 
Our work belongs to adversarial training method and we focus on a observed drawback of adversarial training, that is the learned feature is not well separated in feature space and thus lead to bad generalization.
% The goal of Projected gradient descent(PGD)~\cite{madry2017towards, xu2020adversarial} is to find the most misleading examples located in a small area around the natural examples for the classifier. It calculated the perturbation through maximizing the loss value in the $l_p$ norm ball around the natural data in an iterative manner. PGD perturbs the natural data for a fixed number of gradient step with small step size. After each step, PGD projects the adversarial data back onto the l-p norm ball of the natural data. As a countermeasure, PGD adversarial training trains on adversarial examples calculated by PGD.
% Here, we denote the robust network to be learned as $f_{\theta}$, $\epsilon$ is the perturbation budget, $\mathcal{B_{\epsilon}}$ is a small area around the natural sample $x$ under the given perturbation budget $\epsilon$ and $\mathcal{L}$ is the cross entropy loss function. This inner maximization problem is solved by PGD. Adversarial training has many effective variants.
% TRADES~\cite{zhang2019theoretically} optimizeds an upper bound of adversarial risk; MART~\cite{wang2019improving} found that mis-classified (natural) samples have negligible impact on adversarial robustness hence should pay more attention on those samples in the training process, thus it applies an differentiation of misclassified samples as a regularizer for adversarial risk; AWP~\cite{wu2020adversarial} aims to flatten the weight loss landscape with a designed weight perturbation in the training process to help robust generalization.

\noindent\textbf{Self-supervised Learning} We mention that self-supervised learning techniques, such as Contrastive Learning methods~\cite{chen2020simple} as another main category of related studies to this work. It is because these self-supervised learning techniques are also devised to extract high-quality representations, even in scenarios without any label information~\cite{kolesnikov2019revisiting,jing2020self}. By designing various pretext learning tasks, such as predicting the position of image patches~\cite{doersch2015unsupervised} or classifying image rotation angles~\cite{gidaris2018unsupervised}, self-supervised learning techniques guide the representation learning process on unlabeled examples in a supervised way. Particularly, some recent works also incorporate self-supervised learning techniques into supervised learning tasks \cite{khosla2020supervised, yu2020learning}. However, these techniques cannot be directly adopted into the adversarial learning setting. 

%As data examples can be easily collected in practice while obtaining their corresponding annotations could be very costly, self-supervised learning techniques have been widely adopted for extracting effective representations from unlabeled data samples~\cite{kolesnikov2019revisiting,jing2020self}. By designing various pretext learning tasks, such as predicting position of image patches~\cite{doersch2015unsupervised} or classifying image rotation angles~\cite{gidaris2018unsupervised}, self-supervised learning techniques guide the representation learning process on unlabeled examples in a supervised way. Recent work shows that using additional unlabeled data can help improve model robustness~\cite{carmon2019unlabeled} and self-supervised learning techniques also play positive roles in this process. For example, SS-OOD~\cite{hendrycks2019using} augments the adversarial examples with different rotation angles and utilizes an auxiliary head network to classify rotated adversarial examples. By incorporating this self-supervised pretext learning task into the final objective function, the model can be trained more robust. Moreover, RoCL~\cite{Kim2020} and ACL~\cite{Jiang2020} adversarially train self-supervised models without any labeled data to obtain robust representations. 

\noindent\textbf{Self-supervised Learning and Contrastive Learning} Recent work shows that using additional unlabeled data could help model obtain robust representations; Another categorized of adversarial training variants are proposed to enhance adversarial training proformance with self-supervised learning. For example, SS-OOD ~\cite{hendrycks2019using} arguments the adversarial data with different rotation angles. An auxiliary head network is added to the network to predict the angle and this self-supervised loss is utilized to help robust training. 
% Besides adversarial training with self-supervised assistant . RoCL~\cite{Kim2020} and ACL~\cite{Jiang2020} adversarially train self-supervised models without any labeled data to obtain robust representations. Recent work shows that using additional unlabeled data could help model obtain robust representations; Another categorized of adversarial training variants are proposed to enhance adversarial training proformance with self-supervised learning. For example, SS-OOD ~\cite{hendrycks2019using} arguments the adversarial data with different rotation angles. An auxiliary head network is added to the network to predict the angle and this self-supervised loss is utilized to help robust training. 
% Besides adversarial training with self-supervised assistant . RoCL~\cite{Kim2020} and ACL~\cite{Jiang2020} adversarially train self-supervised models without any labeled data to obtain robust representations. 

\section{Conclusion}
\label{sec:con}
In this work, we propose a novel Adversarial Training framework with Feature Separability (ATFS) to enhance adversarial training. Compared to previous adversarial training methods, ATFS utilizes ATG to capture various types of relations among adversarial and clean examples in adversarial training. As a result, ATFS learns better features for both clean and adversarial samples. Meanwhile, ATFS can achieve both better clean and robust accuracy compared to numerous representative baselines. We conducted comprehensive experiments to show the effectiveness of ATFS and try to understand the reason of ATFS could achieve good performance. In the future, we plan to try if ATFS can be applied to solve other problems in adversarial training, for example, data bias and adversarial training towards multiple bounds attack.

\bibliographystyle{unsrt}
\bibliography{sample}

\clearpage
\appendix

% \large{\bf{Appendix for Graph Neural Networks}}
% \large{\bf{Supplementary Material}}
%\large{\bf{Appendix}}
\onecolumn
\large{\bf{Supplementary Material}}

\section{ATFS Algorithm}
The detailed training algorithm for ATFS is shown in Algorithm~\ref{alg:ATFS}. Specifically, we first construct an ATG $\cG = \{\cV, \cE^+, \cE^-\}$ based on the full training set (Step 3). In each iteration of the training, we first sample a mini-batch $\mathcal{B}$ (Step 6), and then generate adversarial samples using Projected Gradient Descent (PGD) for samples in the current batch $\mathcal{B}$  (Step 7). Based on the clean samples $\{\vx_i\}$ and their corresponding adversarial samples $\{\vx'_i\}$ in this mini-batch, we extract the subgraph $\mathcal{G}_B$ from $\cG$ which contains clean and adversarial samples as nodes and their corresponding links (Step 8). Giving the subgraph $\mathcal{G}_B$, we can easily calculate its Feature Separability Loss and we minimize the objective in Eq.~\eqref{objective} with respect to the model parameters $\theta$ by gradient descent. The procedures from Step 6 to Step 10 will be repeated until convergence. 

\begin{algorithm}[h]
\begin{algorithmic}[1]
%\setstretch{1.2}
   \STATE { \textbf{Input:}} Training data $\cD = \{(\vx_i, y_i) \}^{n}_{i=1}$, learning rate $\gamma$, momentum $m$, weight decay $w$, number of epochs $T$, batch size $b$
   \STATE { \textbf{Output:}} An adversarially robust model $f_{\mathbf{\theta}}$ 
   \STATE Construct an ATG $\cG = \{\cV, \cE^+, \cE^-\}$;
  \FOR{epoch $= 1$, $\dots$, $T$}
    \FOR {mini-batch = $1, \dots, \left \lceil{\frac{n}{b}}\right \rceil $}
        \STATE Sample a mini-batch $\mathcal{B}=\{(\vx_i, y_i) \}^{b}_{i=1}$ from $\cD$;
        \STATE Generate adversarial samples \{$x'_i\}$ for $\mathcal{B}$ by PGD attack;
        \STATE Extract a subgraph from ATG for the mini-batch 
        $\mathcal{\cG_{\cB}} = \{\cV_\cB, \cE_\cB^+, \cE_\cB^- \}$;
    \STATE Update $\theta$ based on the objective function in Eq.~\eqref{objective} by gradient descent with momentum $m$ and weight decay $w$;
  \ENDFOR
 \ENDFOR
    \caption{Adversarial Training with Feature Separability}
\label{alg:ATFS}

\end{algorithmic}
\end{algorithm}

\section{Experiment Result on WRN-34}

\begin{table*}[h]
\caption{Performance Comparison on CIFAR10 with WRN-34-10. }
\label{tab: CIFAR10_WRN34}
\centering
\begin{tabular}{@{}l|c|c|c|c|c@{}}
\hline
\hline
       & Clean & FGSM & PGD-20(8/255)  & $\text{CW}_{\inf}$(8/255) & AA(8/255)    \\
\hline       
AT        & 87.20 &   63.47±0.07   &  46.84±0.03 & 47.48±0.03 & 44.04 \\
%RoCL+TRADES             & 84.55 & 43.85   & 14.29 & - & -\\
%RoCL+AT+SS              & - & 49.66   & 14.44    & - & - \\
%MLA & 86.61 &      &   &  & 47.41\\

TRADES ($1/\lambda=5.0$)   & 85.23 & 66.19±0.03    & 53.87±0.00 & 52.50±0.03   &  52.40 \\
TRADES ($1/\lambda=6.0$)   & 84.92 &  66.85±0.04   &  55.34±0.03   & 53.81±0.03 & 53.08 \\
SS-OOD  & 81.61 & 62.49±0.00 & 49.80±0.00& 49.27±0.00 & 48.88  \\ 
MART & 83.62 &    67.15±0.03   &   56.35±0.00    & 52.69±0.03   & 51.69 \\
%AWP & 85.36 &   69.25±0.00   & 59.24±0.03 & 57.10±0.00 & 56.33   \\
%GAIRAT & 85.75 & & 57.81±0.54 & &\\
\hline
%clean+CL($\alpha/\beta= 100$,$\eta$ = 1)     & \textbf{85.89} & 50.93±0.004  & 15.89   &    48.97      & 12.22    \\
%ATFS(clean)     & 85.91 & 67.24±0.00  & 54.52±0.04   & 55.02±0.08  &  47.91±0.04    \\
AT + ATFS & 85.27 & 68.00±0.00  & 56.00±0.00  &  51.10±0.00 &   51.60   \\
%ATFS(clean) + AWP & 85.22 &  68.32±0.03   &  57.24±0.03 & 56.66±0.03 &  49.62±0.04\\
%AWP + ATFS & 85.54 &   68.35±0.03  &    58.63±0.03         & 50.26±0.00 & 57.84 \\
TRADES + ATFS & 85.17 & 66.68±0.05  & 56.24±0.00 & 54.01±0.01 & 53.42 \\
MART + ATFS & 84.64 & 68.3±0.00 & 58.78±0.00 & 53.25±0.00 & 52.69\\

\hline
\hline
\end{tabular}

\end{table*}

\end{document}